\documentclass{article}

\usepackage{spconf,graphicx} 

\usepackage{amsmath,amsfonts}
\usepackage{algorithmic}
\usepackage{algorithm}
\usepackage{array}
\usepackage{booktabs}
\usepackage{textcomp}
\usepackage{xcolor}
\usepackage{stfloats}
\usepackage{url}
\usepackage{verbatim}
\usepackage{dsfont}
\usepackage{multicol}
\usepackage{multirow}
\usepackage{blindtext}
\usepackage{caption}
\usepackage{subcaption}

\usepackage{fancyhdr}
\fancypagestyle{arxiv}{
  \fancyhf{} 
  \fancyhead[C]{\footnotesize Accepted at 2026 IEEE International Conference on Acoustics, Speech and Signal Processing (ICASSP), May 4--8, 2026, Barcelona, Spain}
}

\title{Beyond Amplitude: Channel State Information Phase-Aware Deep Fusion for Robotic Activity Recognition}

\name{Rojin Zandi $^\dagger$, Hojjat Salehinejad {$^{\ddagger\star}$}, Senior Member, IEEE, \textit{and Milad Siami$^\dagger$, Senior Member, IEEE}
 \thanks{This material is based upon work supported in part by the U.S. Office of Naval Research under Grant Award N00014-21-1-2431; and in part by the U.S. National Science Foundation under Grant Award 2208182.}
}

\address{$^\dagger$Department of Electrical \& Computer Engineering, Northeastern University, Boston, MA, USA\\
$^\ddagger$Kern Center for the Science of Health Care Delivery, Mayo Clinic, Rochester, MN, USA\\
$^\star$Department of Artificial Intelligence and Informatics, Mayo Clinic, Rochester, MN, USA}

\begin{document}

\maketitle 
\thispagestyle{arxiv} 

\begin{abstract}

Wi-Fi Channel State Information (CSI) has emerged as a promising non-line-of-sight sensing modality for human and robotic activity recognition. However, prior work has predominantly relied on CSI amplitude while underutilizing phase information, particularly in robotic arm activity recognition. In this paper, we present GateFusion-Bidirectional Long Short-Term Memory network (GF-BiLSTM) for WiFi sensing in robotic activity recognition. GF-BiLSTM is a two-stream gated fusion network that encodes amplitude and phase separately and adaptively integrates per-time features through a learned gating mechanism. We systematically evaluate state-of-the-art deep learning models under a Leave-One-Velocity-Out (LOVO) protocol across four input configurations: amplitude only, phase only, amplitude + unwrapped phase, and amplitude + sanitized phase. Experimental results demonstrate that incorporating phase alongside amplitude consistently improves recognition accuracy and cross-speed robustness, with GF-BiLSTM achieving the best performance. To the best of our knowledge, this work provides the first systematic exploration of CSI phase for robotic activity recognition, establishing its critical role in Wi-Fi–based sensing.

\end{abstract}
\section{Introduction}
\label{sec:intro}
Robotic systems are increasingly deployed across diverse domains, making automatic recognition of their activities essential for safe and reliable operation \cite{javaid2021substantial}. Conventional solutions typically rely on cameras or Light Detection and Ranging (LiDAR) systems to monitor robot movements and arm activities, but these sensors require a clear line-of-sight and may raise privacy concerns \cite{petrlik2021lidar, correa2012mobile, yang2024mm}. This has motivated the exploration of alternative non-visual sensing modalities \cite{zandi2023robot}.

Wi-Fi sensing has recently emerged as a promising approach for human and robotic activity recognition \cite{10890231,10446641, salehinejad2023joint}. Channel State Information (CSI) extracted from Wi-Fi signals contains both amplitude and phase components: the amplitude reflects how much of the transmitted signal reaches the receiver after environmental distortion, while the phase captures fine-grained path differences caused by reflections, scattering, and multipath propagation. Most prior work on Wi-Fi–based activity recognition has focused on CSI amplitude while disregarding phase, since raw phase measurements are corrupted by hardware-induced timing and frequency offsets \cite{yousefi2017surv}. Recent studies demonstrate that, once calibrated, CSI phase combined with amplitude can improve human activity recognition accuracy \cite{quy2025enhanced}. However, the role of CSI phase in robotic activity recognition remains largely unexplored.

This work studies how to incorporate CSI phase information, including phase only, amplitude only, amplitude + unwrapped phase, and amplitude + sanitized phase, and introduces GateFusion-Bidirectional Long Short-Term Memory networks (GF-BiLSTM), a two-stream gated-fusion BiLSTM that encodes amplitude and phase separately before temporal fusion. The impact of these input configurations is evaluated by training and testing several established deep learning models, including Convolutional Neural Networks (CNN) \cite{lecun1998gradient},  LSTM \cite{hochreiter1997long}, BiLSTM \cite{schuster1997bidirectional}, Vision Transformer (ViT) \cite{dosovitskiy2020image}, and the Bidirectional ViT Concatenated (BiVTC) model \cite{zandi2024robofisense}. Experiments are conducted on the RoboFiSense benchmark dataset~\cite{zandi2024robofisense}, which contains eight robotic arm actions performed at low, medium, and high velocities. To evaluate generalization across different motion speeds, we adopt a Leave-One-Velocity-Out (LOVO) protocol, where models are trained on two velocities and tested on the unseen third.

This study contributes a systematic assessment of CSI phase for robotic-arm action recognition, an architecture tailored to exploit it, and a rigorous cross-speed evaluation. First, across strong baselines, amplitude+phase consistently outperforms single-modality inputs, establishing phase as a complementary cue rather than a substitute. Second, the proposed GF-BiLSTM achieves the best performance by learning per-stream temporal representations and fusing them adaptively at each time step. Third, under the LOVO, GF-BiLSTM markedly improves generalization to unseen execution speeds, addressing a key robustness gap in Wi-Fi sensing. Taken together, these results provide the first targeted evidence that calibrated phase meaningfully boosts accuracy and cross-speed robustness in robotic-arm activity recognition.

\begin{figure*}
\footnotesize
    \centering
    \includegraphics[width=1\linewidth]{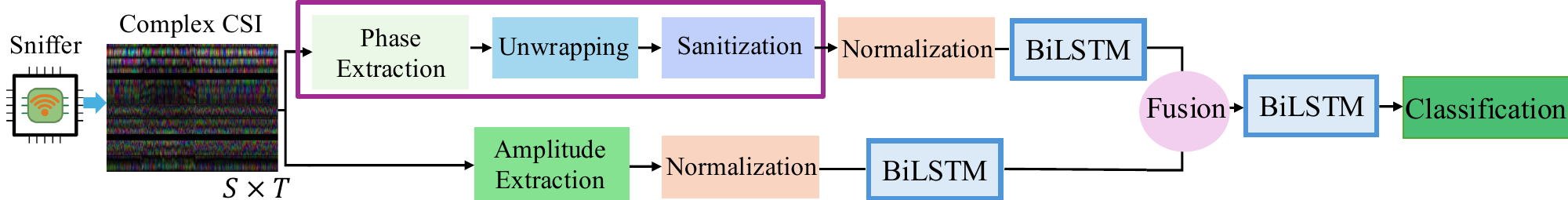}
    \caption{Complex CSI is split into amplitude and phase. The phase branch performs temporal unwrapping and packet-wise linear sanitization (purple box), then per-time layer normalization; the amplitude branch is normalized identically. Each stream is encoded by a 1-layer BiLSTM, their per-time features are fused by a learned gate (fusion), and a deeper BiLSTM with a final classifier produces action labels.}
    \label{fig:flow}
\end{figure*}
\begin{figure*}[t]
\centering
\begin{subfigure}[b]{0.25\textwidth}
  \includegraphics[width=\linewidth]{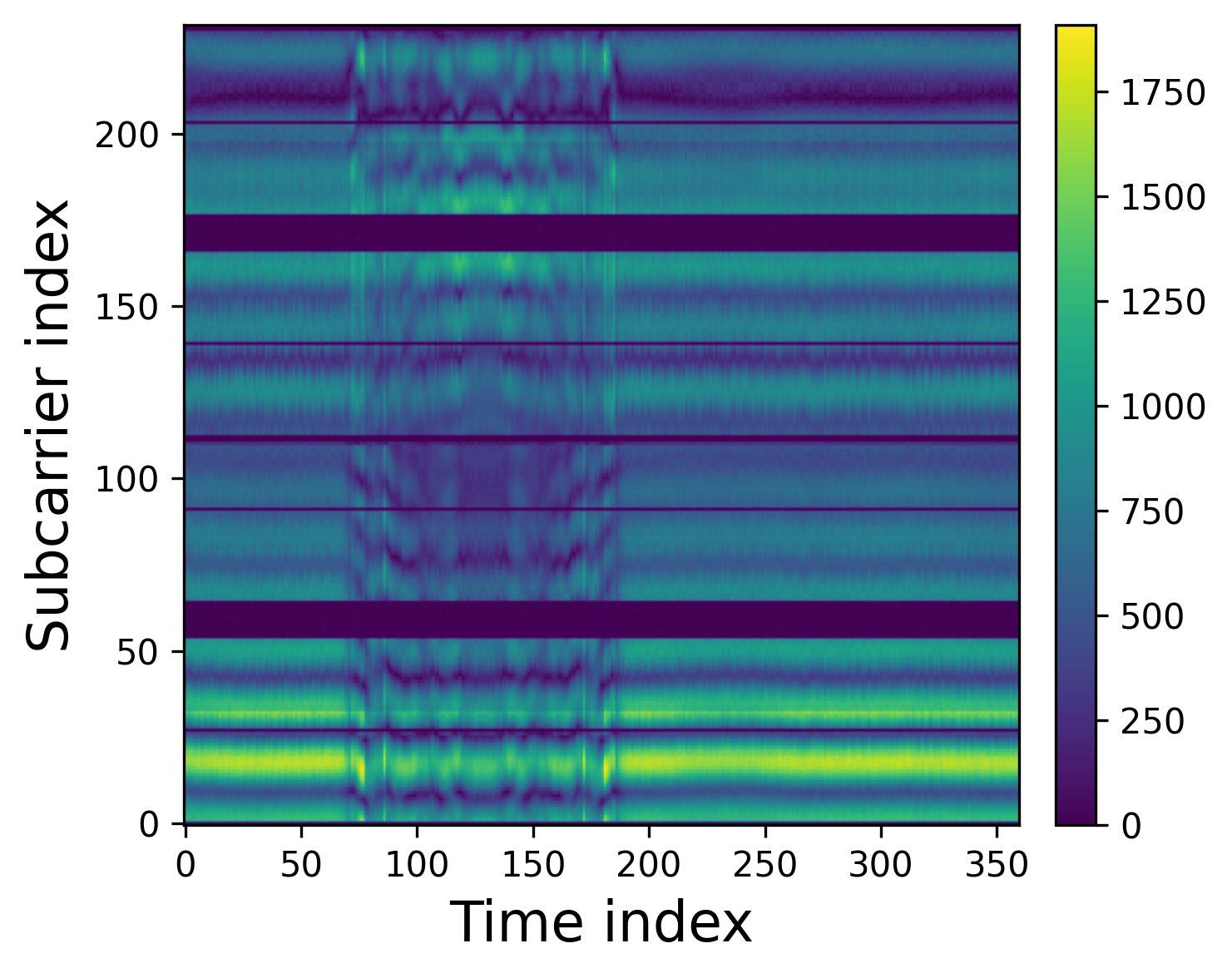}
  \caption{Amplitude heatmap} \label{amp-heat}
\end{subfigure}\hfill
\begin{subfigure}[b]{0.25\textwidth}
  \includegraphics[width=\linewidth]{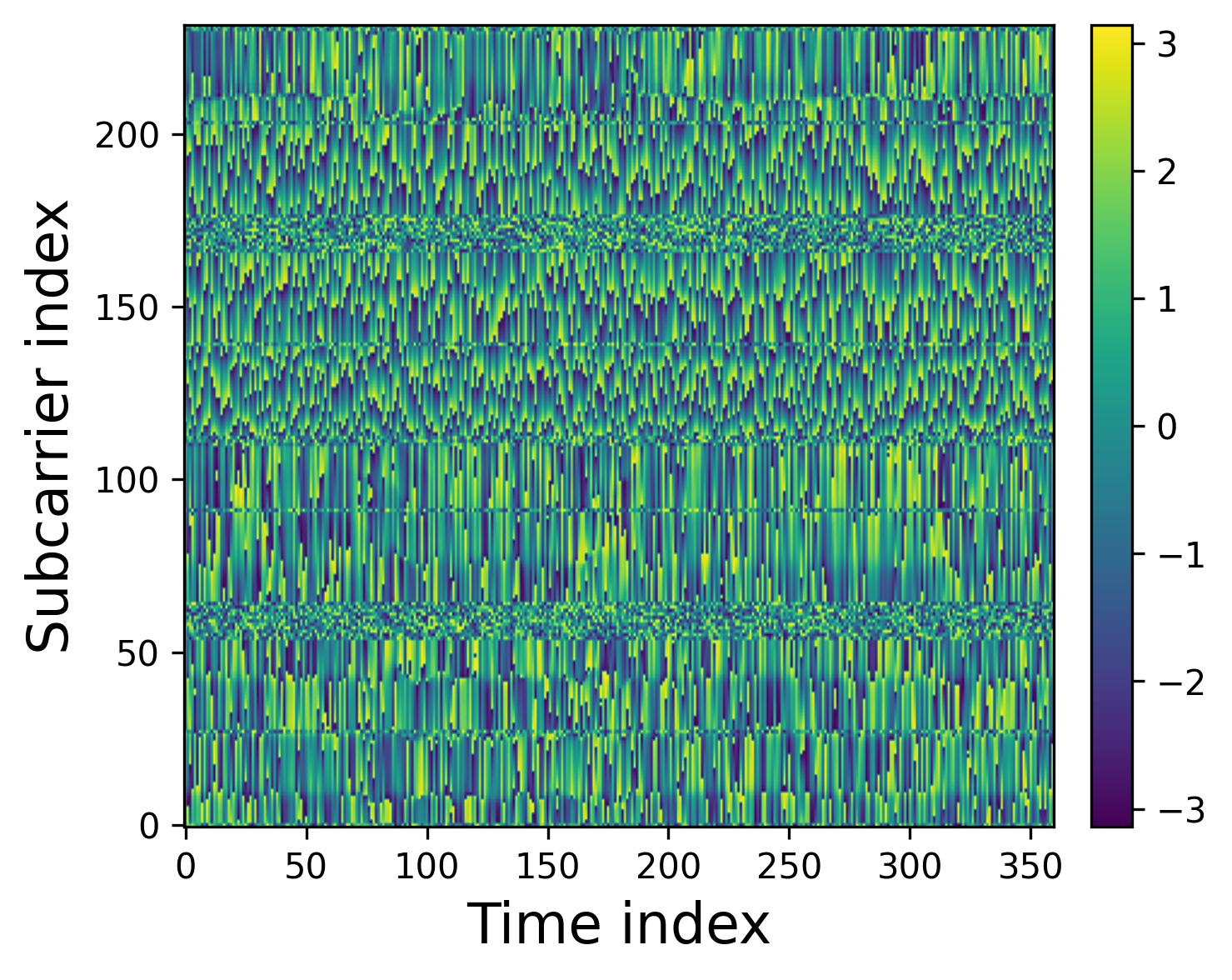}
  \caption{Raw phase heatmap}\label{raw-ph-heat}
\end{subfigure}\hfill
\begin{subfigure}[b]{0.25\textwidth}
  \includegraphics[width=\linewidth]{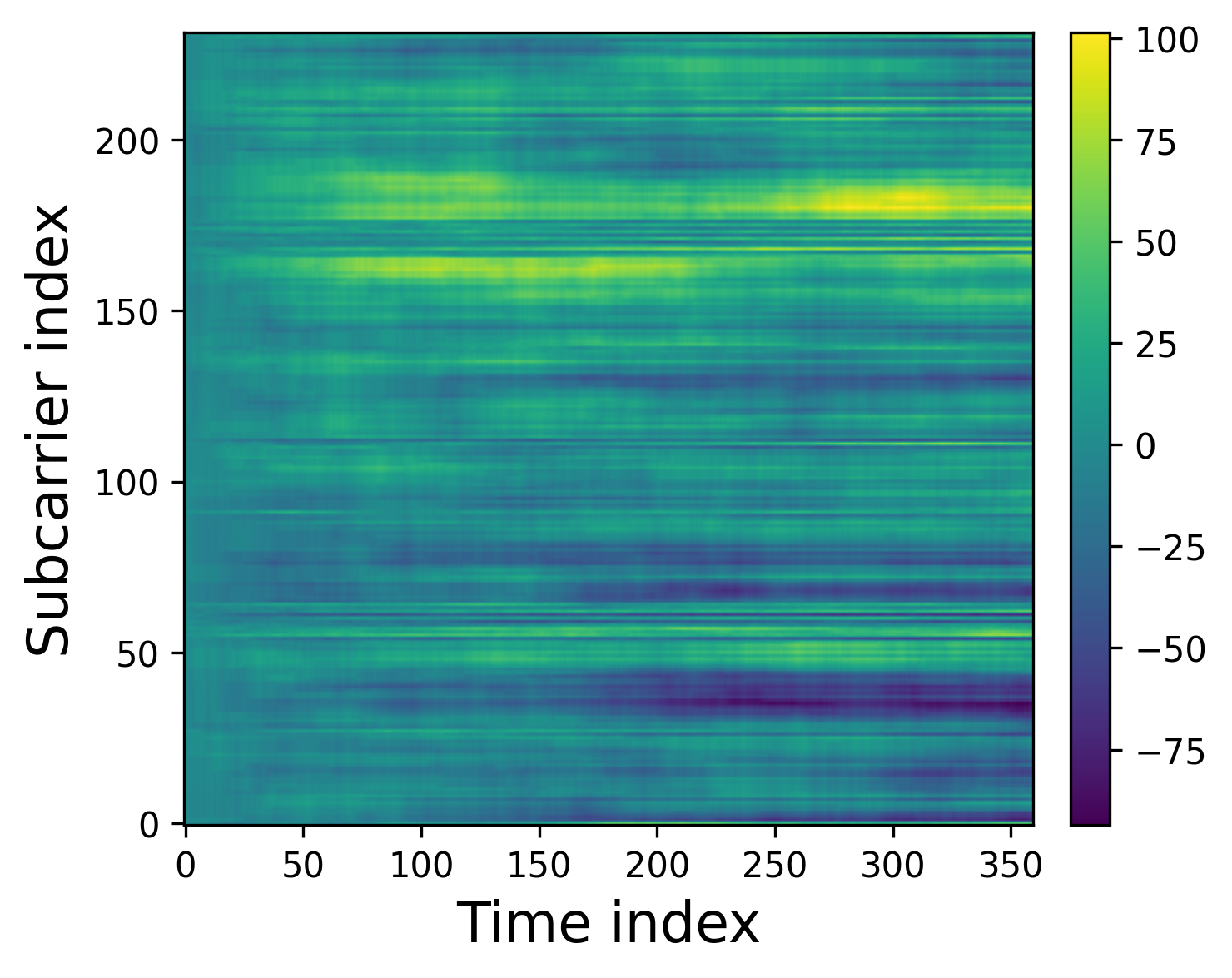}
  \caption{Unwrapped phase heatmap}\label{unw-ph-heat}
\end{subfigure}\hfill
\begin{subfigure}[b]{0.25\textwidth}
  \includegraphics[width=\linewidth]{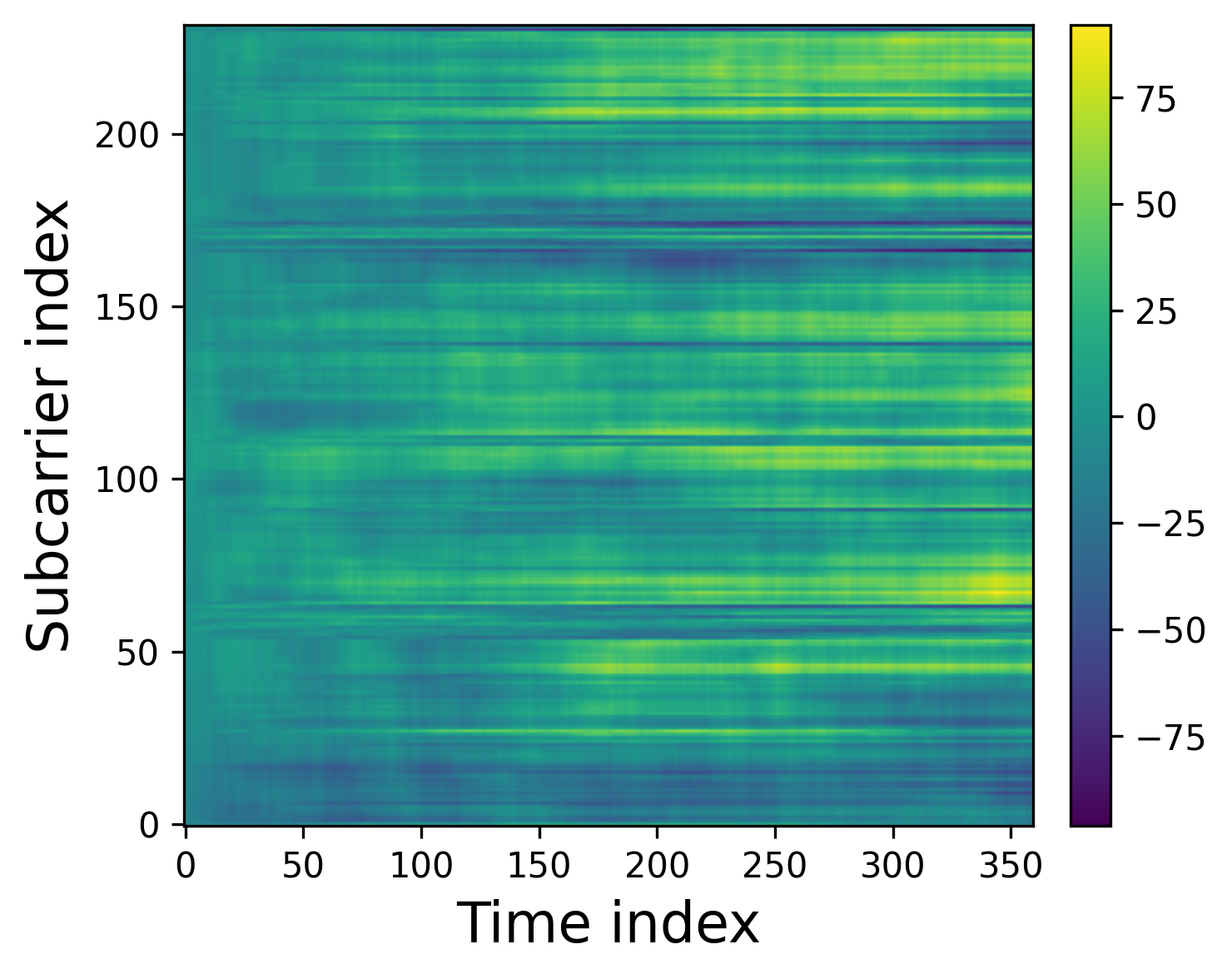}
  \caption{Sanitized phase heatmap}\label{san-ph-heat}
\end{subfigure}
\begin{subfigure}[b]{0.25\textwidth}
  \includegraphics[width=\linewidth]{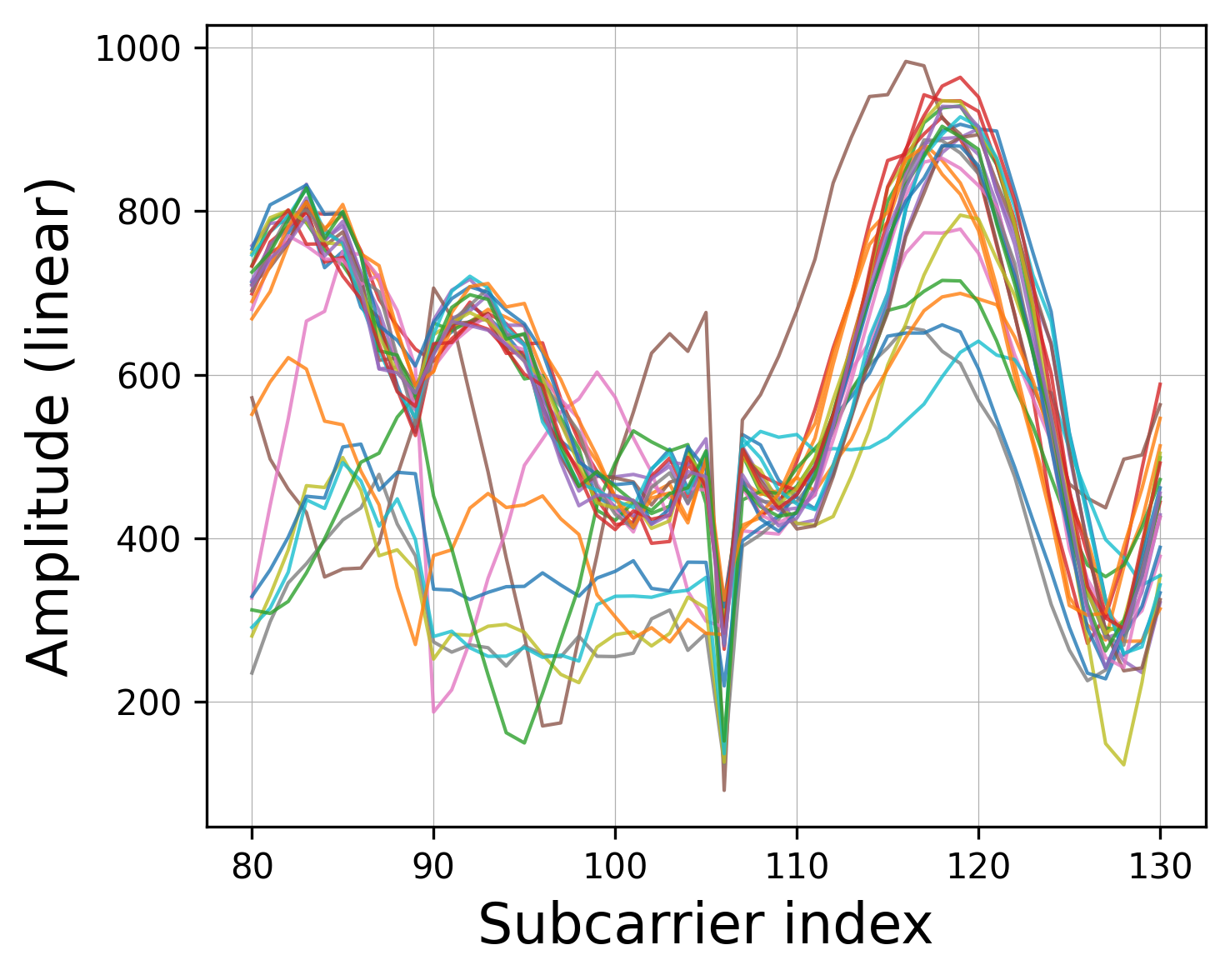}
  \caption{Amplitude overlay} \label{Amp-ov}
\end{subfigure}\hfill
\begin{subfigure}[b]{0.25\textwidth}
  \includegraphics[width=\linewidth]{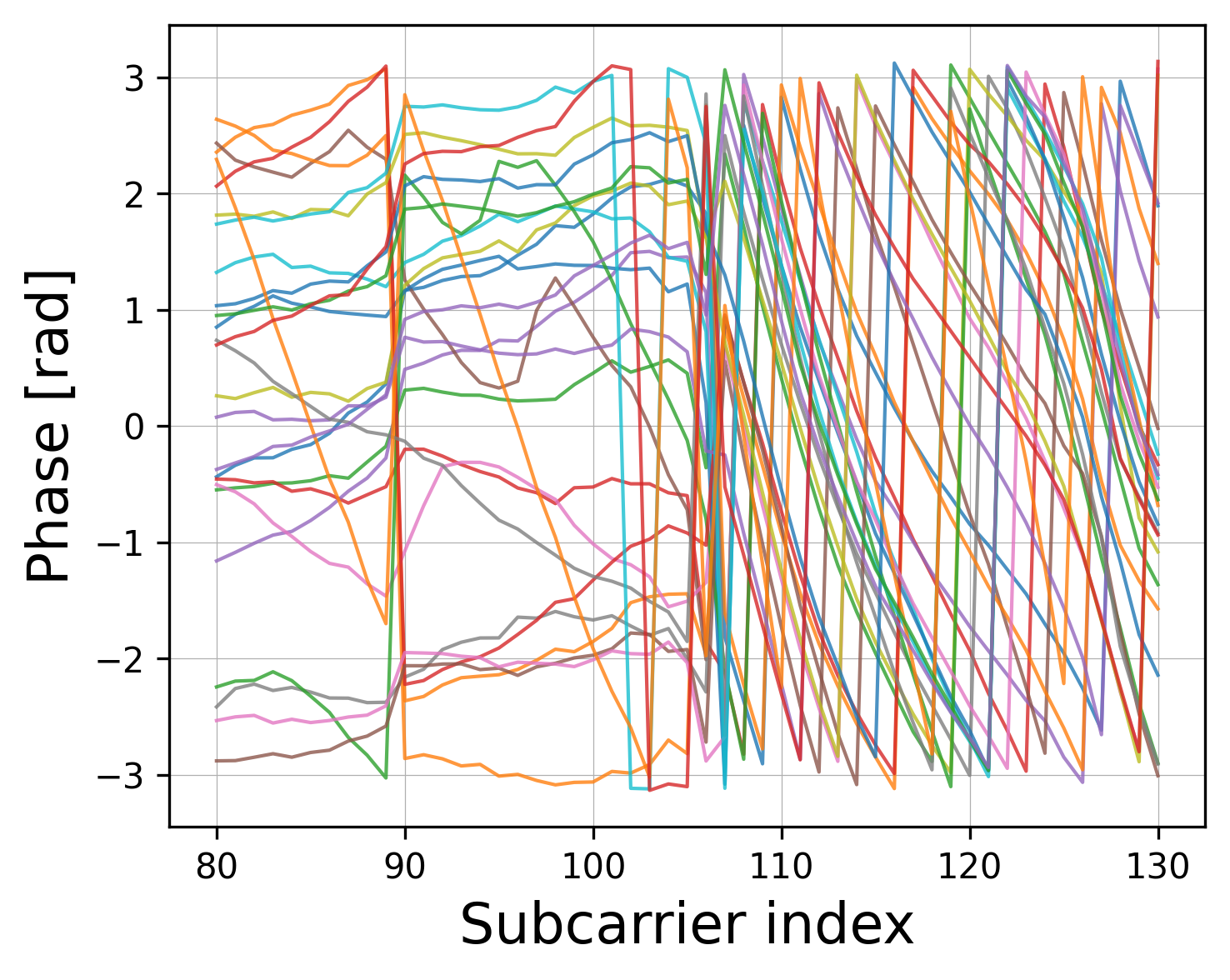}
  \caption{Raw phase overlay}\label{raw-ph-ov}
\end{subfigure}\hfill
\begin{subfigure}[b]{0.25\textwidth}
  \includegraphics[width=\linewidth]{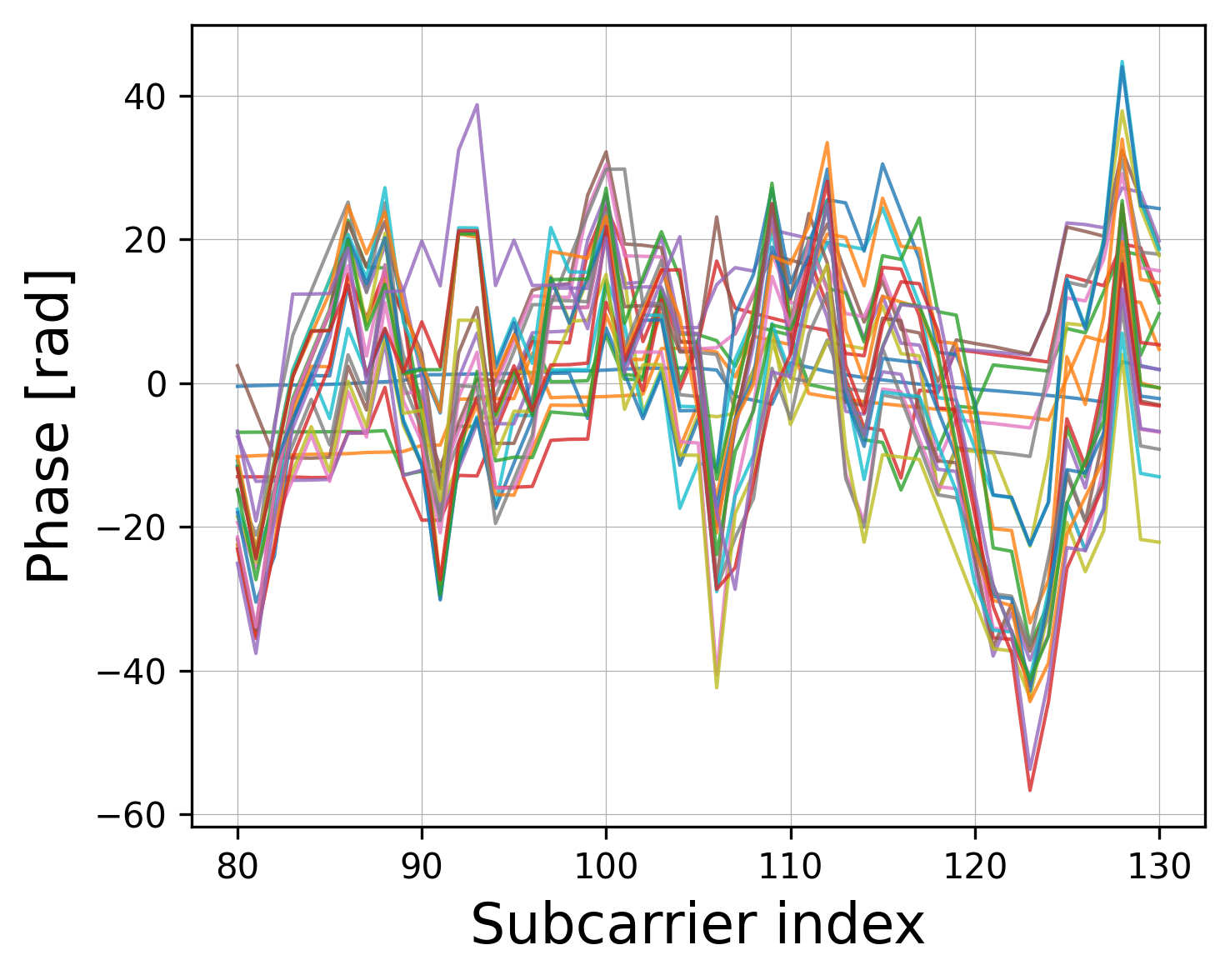}
  \caption{Unwrapped phase overlay}\label{unw-ph-ov}
\end{subfigure}\hfill
\begin{subfigure}[b]{0.25\textwidth}
  \includegraphics[width=\linewidth]{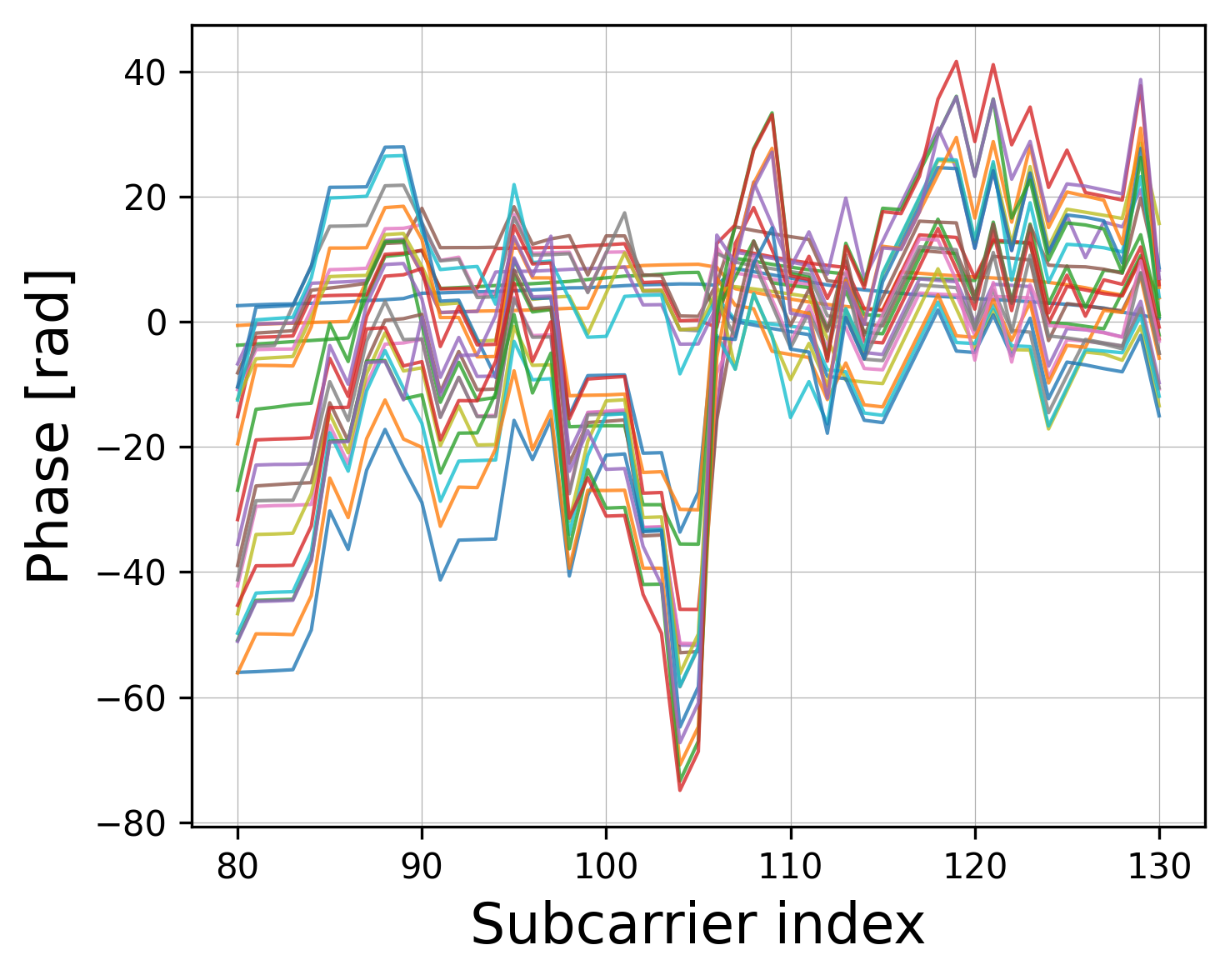}\label{san-ph-ov}
  \caption{Sanitized phase overlay}
\end{subfigure}
\caption{CSI visualization from one sniffer across time and subcarriers. Top row (a–d): heatmaps of amplitude, raw phase, unwrapped phase, and linearly sanitized phase. Bottom row (e–h): corresponding overlay plots, where each curve is a time snapshot across subcarriers. Heatmaps show full temporal evolution, while overlays illustrate frequency response at selected times, highlighting the effect of phase unwrapping and sanitization.}
\label{fig:ph&amp}
\end{figure*}
\vspace{-0.55cm}
\section{Method}
\label{sec:method}
This section outlines the end-to-end pipeline from raw Wi-Fi CSI to action labels (Fig.~\ref{fig:flow}). It briefly introduces the data and notation, summarizes the input preprocessing, presents the proposed GF-BiLSTM architecture, and states the training and evaluation setup.

\subsection{Channel State Information Measurements}
Let $\mathbf{H}\in\mathbb{C}^{S\times T}$ denote CSI over $S$ subcarriers and $T$ timestamps, with entries
\begin{equation}
    h_{k,t} \;=\; a_{k,t}\,e^{j\phi_{k,t}},
\end{equation}
where $a_{k,t}=|h_{k,t}|$ is amplitude and $\phi_{k,t}=\angle h_{k,t}$ is phase. Stack these into
\begin{equation}
    \mathbf{A}=\big[a_{k,t}\big]_{k,t}\in\mathbb{R}^{S\times T}, \qquad
    \boldsymbol{\Phi}=\big[\phi_{k,t}\big]_{k,t}\in\mathbb{R}^{S\times T},
\end{equation}
as presented in Fig. \ref{fig:ph&amp}. Amplitude primarily reflects attenuation, whereas phase captures path-length changes and fine-grained temporal dynamics \cite{wang2015understanding,zandi2024enhancing}. A subcarrier-first, time-second layout is used; thus the model’s time-slice vectors are $\mathbf{a}_t=\mathbf{A}_{:,t}\in\mathbb{R}^{S}$ and $\boldsymbol{\phi}_t=\boldsymbol{\Phi}_{:,t}\in\mathbb{R}^{S}$.
\vspace{-0.3cm}
\subsection{Phase Preprocessing} \label{sec:php}
Generally, the raw phase is corrupted by hardware impairments (carrier/sampling offsets, random packet rotations) \cite{yousefi2017surv}. A physically meaningful trajectory is obtained by temporal unwrapping followed by packet-wise linear sanitization.

\subsubsection{Temporal Unwrapping}
Phases are wrapped to $(-\pi,\pi]$, as shown in Fig. \ref{raw-ph-heat} and \ref{raw-ph-ov}. For each subcarrier $k$, an integer sequence $\{n_{k,t}\}_{t=1}^{T}$ is chosen to minimize discontinuities between consecutive timestamps and
\begin{equation}
    \tilde{\phi}_{k,t} \;=\; \phi_{k,t} + 2\pi\, n_{k,t},
\end{equation}
yielding continuous trajectories; stacking gives $\tilde{\boldsymbol{\Phi}}\in\mathbb{R}^{S\times T}$.

\subsubsection{Linear Sanitization}
Residual synchronization creates an approximately linear trend across subcarriers within a packet. For each time index $t$,
\begin{equation}
    \tilde{\boldsymbol{\phi}}_{t} \;=\; \alpha_t\,\mathbf{k} + \beta_t\,\mathbf{1} + \boldsymbol{\epsilon}_t,
    \qquad
    \tilde{\boldsymbol{\phi}}_{t}=\tilde{\boldsymbol{\Phi}}_{:,t}\in\mathbb{R}^{S},
\end{equation}
with $\mathbf{k}=[1,2,\ldots,S]^\top$, $\mathbf{1}\in\mathbb{R}^{S}$, slope $\alpha_t$, offset $\beta_t$, and residual $\boldsymbol{\epsilon}_t$ \cite{diaz2023channel}. With $\mathbf{X}=[\,\mathbf{k}\ \ \mathbf{1}\,]\in\mathbb{R}^{S\times 2}$ and $\boldsymbol{\theta}_t=[\alpha_t\ \ \beta_t]^\top$, least squares gives
\begin{equation}
    \boldsymbol{\theta}_t^\star = \arg\min_{\boldsymbol{\theta}\in\mathbb{R}^{2}} \|\tilde{\boldsymbol{\phi}}_{t}-\mathbf{X}\boldsymbol{\theta}\|_2^2
    = (\mathbf{X}^\top\mathbf{X})^{-1}\mathbf{X}^\top \tilde{\boldsymbol{\phi}}_{t},
\end{equation}
and the sanitized phase
\begin{equation}
    \hat{\boldsymbol{\phi}}_{t} \;=\; \tilde{\boldsymbol{\phi}}_{t} - \mathbf{X}\boldsymbol{\theta}_t^\star,
    \qquad
    \hat{\boldsymbol{\Phi}}=\big[\hat{\boldsymbol{\phi}}_{t}\big]_{t=1}^{T}\in\mathbb{R}^{S\times T}.
\end{equation}
Unless stated otherwise, the model uses $\boldsymbol{\phi}_t=\hat{\boldsymbol{\Phi}}_{:,t}$; ablations with unwrapped $\tilde{\boldsymbol{\Phi}}$ are also reported.
\vspace{-0.3cm}
\subsection{Gate Fusion-BiLSTM}
Let $h$ denote the per-direction hidden width of each LSTM and $C$ the number of classes. Two temporal streams process amplitude and phase separately and are fused after one feature-extraction layer, followed by deeper temporal modeling and classification.

\textit{Per-time normalization:}
Layer normalization across subcarriers is applied for each stream and time index:
\begin{align}
\tilde{\mathbf{a}}_t
&= \mathrm{LN}_A(\mathbf{a}_t)
= \boldsymbol{\gamma}^{A}\odot\frac{\mathbf{a}_t-\mu^{A}_t\mathbf{1}}{\sqrt{(\sigma^{A}_t)^2+\varepsilon}}
+ \boldsymbol{\beta}^{A}, \\
\tilde{\boldsymbol{\phi}}_t
&= \mathrm{LN}_P(\boldsymbol{\phi}_t)
= \boldsymbol{\gamma}^{P}\odot\frac{\boldsymbol{\phi}_t-\mu^{P}_t\mathbf{1}}{\sqrt{(\sigma^{P}_t)^2+\varepsilon}}
+ \boldsymbol{\beta}^{P},
\end{align}
with learned $\boldsymbol{\gamma}^{A},\boldsymbol{\beta}^{A},\boldsymbol{\gamma}^{P},\boldsymbol{\beta}^{P}\in\mathbb{R}^{S}$, means $\mu^{A}_t=\tfrac{1}{S}\mathbf{1}^\top \mathbf{a}_t$ and $\mu^{P}_t=\tfrac{1}{S}\mathbf{1}^\top \boldsymbol{\phi}_t$, and variances $(\sigma^{A}_t)^2,(\sigma^{P}_t)^2$ defined analogously. During training only, an optional modality-dropout masks one stream per sample with a small probability, using masks that are constant across time. There is no masking at inference.

\textit{First-level feature extraction:}
Separate bidirectional LSTMs encode amplitude and phase. Let $\mathbf{h}^{A,\rightarrow}_t,\mathbf{c}^{A,\rightarrow}_t\in\mathbb{R}^{h}$ and $\mathbf{h}^{A,\leftarrow}_t,\mathbf{c}^{A,\leftarrow}_t\in\mathbb{R}^{h}$ denote hidden and cell states of the forward and backward amplitude LSTMs at time $t$ (standard LSTM recurrences). The phase stream defines $\mathbf{h}^{P,\rightarrow}_t,\mathbf{c}^{P,\rightarrow}_t$ and $\mathbf{h}^{P,\leftarrow}_t,\mathbf{c}^{P,\leftarrow}_t$ analogously. Concatenating directions and projecting to width $h$ yields
\begin{align}
\mathbf{u}^{A}_t &= \rho\!\big(\mathbf{W}^{A}[\mathbf{h}^{A,\rightarrow}_t;\mathbf{h}^{A,\leftarrow}_t]+\mathbf{b}^{A}\big)\in\mathbb{R}^{h}, \\
\mathbf{u}^{P}_t &= \rho\!\big(\mathbf{W}^{P}[\mathbf{h}^{P,\rightarrow}_t;\mathbf{h}^{P,\leftarrow}_t]+\mathbf{b}^{P}\big)\in\mathbb{R}^{h},
\end{align}
where $\rho$ is ReLU, $[\,\cdot\,;\,\cdot\,]$ denotes concatenation along the feature dimension, and $\mathbf{W}^{A},\mathbf{W}^{P}\in\mathbb{R}^{h\times 2h}$ with biases $\mathbf{b}^{A},\mathbf{b}^{P}\in\mathbb{R}^{h}$.

\textit{Fusion:}
Per-time gated fusion combines the two streams:
\begin{align}
\mathbf{g}_t &= \sigma\!\big(\mathbf{W}^{g}[\mathbf{u}^{A}_t;\mathbf{u}^{P}_t]+\mathbf{b}^{g}\big)\in(0,1)^{h}, \\
\mathbf{z}_t &= \mathbf{g}_t\odot \mathbf{u}^{A}_t + (\mathbf{1}-\mathbf{g}_t)\odot \mathbf{u}^{P}_t \in \mathbb{R}^{h},
\end{align}
where $\sigma$ is the logistic sigmoid, $\odot$ the Hadamard product, $\mathbf{1}\in\mathbb{R}^{h}$ the all-ones vector, and $\mathbf{W}^{g}\in\mathbb{R}^{h\times 2h}$ with bias $\mathbf{b}^{g}\in\mathbb{R}^{h}$.

\textit{Deeper temporal modeling, pooling, and classification:}
The fused sequence $(\mathbf{z}_1,\ldots,\mathbf{z}_T)$ is processed by a deeper bidirectional LSTM with hidden states $\mathbf{h}^{\rightarrow}_t,\mathbf{h}^{\leftarrow}_t\in\mathbb{R}^{h}$ and output $\mathbf{h}_t=[\mathbf{h}^{\rightarrow}_t;\mathbf{h}^{\leftarrow}_t]\in\mathbb{R}^{2h}$. Temporal average pooling produces $\bar{\mathbf{h}}=\tfrac{1}{T}\sum_{t=1}^{T}\mathbf{h}_t\in\mathbb{R}^{2h}$, followed by a two-layer head
\begin{equation}
\mathbf{q}=\rho(\mathbf{W}_1\bar{\mathbf{h}}+\mathbf{b}_1)\in\mathbb{R}^{2h}, \qquad
\boldsymbol{\ell}=\mathbf{W}_2\mathbf{q}+\mathbf{b}_2\in\mathbb{R}^{C},
\end{equation}
with $\mathbf{W}_1\in\mathbb{R}^{2h\times 2h}$, $\mathbf{W}_2\in\mathbb{R}^{C\times 2h}$ and biases $\mathbf{b}_1,\mathbf{b}_2$. The predicted distribution is $\mathbf{p}=\mathrm{softmax}(\boldsymbol{\ell})$, and training minimizes cross-entropy $\mathcal{L}=-\log p_y$ for label $y\in\{1,\ldots,C\}$.

\vspace{-0.3cm}

\subsection{Data Representation for Other Models}
The training of the other models, used in this paper, proceeds with the phase preprocessing explained in Sec. \ref{sec:php}, which is shown by the purple box in Fig. \ref{fig:flow}, plus amplitude extraction. After preprocessing, each CSI measurement consists of an amplitude component $\mathbf{A} \in \mathbb{R}^{S\times T}$ and a corresponding phase component $\phi$. Depending on the task, these can be used individually or combined as multi-channel inputs. In the following experiments, we consider both single-modality and joint representations of amplitude and phase to assess their contribution to robotic activity recognition.
\vspace{-0.35cm}
\section{Experiments}\label{sec:experiments}
\vspace{-0.35cm}
\subsection{Dataset}\vspace{-0.1cm}
We evaluate our approach on the RoboFiSense dataset~\cite{zandi2024robofisense}, which records the CSI of a Franka Emika robotic arm performing eight distinct activities (Arc, Elbow, Rectangle, Silence, SLFW, SLRL, SLUD, Triangle). Each activity is executed under three velocities: low ($\mathcal{V}_1$), medium ($\mathcal{V}_2$), and high ($\mathcal{V}_3$). To assess robustness against motion-speed variations, we adopt the LOVO evaluation protocol: models are trained on two velocity subsets and tested on the remaining unseen velocity. This setup provides a stringent measure of generalization, as models must classify activities under execution speeds not observed during training.  

    

\vspace{-0.3cm}
\subsection{Input Configurations}\label{sec:input-con}
Four CSI representations are evaluated to isolate the role of phase and its preprocessing:
\begin{enumerate}
    \item \textit{Phase only:} unwrapped phase without sanitization; denoted $\tilde{\boldsymbol{\Phi}}$ with shape $S\times T$, captures fine timing/path-length cues but retains packet-wise bias and jitter, yielding low SNR when used alone.
    \item \textit{Amplitude only:} magnitude of CSI; denoted $\mathbf{A}$ with shape $S\times T$, provides a stable, high-SNR baseline reflecting path loss and multipath; inexpensive but discards timing cues.
    \item \textit{Amplitude+Phase (Unwrapped):} two-channel stack $[\,\mathbf{A};\tilde{\boldsymbol{\Phi}}\,]$ with shape $2\times S\times T$, combines amplitude’s stability with phase’s temporal detail while avoiding linear fits; works well with GF-BiLSTM since the gate can down-weight noisy phase windows.
    \item \textit{Amplitude+Phase (Sanitized):} two-channel stack $[\,\mathbf{A};\hat{\boldsymbol{\Phi}}\,]$ with shape $2\times S\times T$, uses per-packet linear trend removal to reduce drift/bias, yielding a small but consistent accuracy gain at higher preprocessing cost.
\end{enumerate}
All two-channel inputs follow a modality$\times$subcarrier$\times$time layout. Table~\ref{tab:results} summarizes results under these settings.
\vspace{-0.3cm}
\subsection{Training Setup}
The GF-BiLSTM uses one BiLSTM layer per stream (amplitude/phase) and a post-fusion stack of two BiLSTM layers with hidden width $h{=}128$ per direction, followed by temporal averaging and a two-layer MLP head. Optimization employs AdamW (learning rate $1{\times}10^{-4}$, weight decay $2{\times}10^{-5}$) with cross-entropy loss. Regularization includes dropout $0.2$ in recurrent/linear layers and stream-level modality dropout $0.05$ (training only). Batch size is $8$.

Baselines (CNN, LSTM, BiLSTM, ViT, BiVTC) are trained under the same data splits and optimizer settings with identical dropout and batch size. For two-channel inputs, the first-layer channel dimension is doubled while keeping the overall capacity comparable. Early stopping on validation performance is applied to all models.

\begin{table}[!t]
\centering
\footnotesize
\caption{Accuracy of the models for the leave-one-velocity-out cross-validation study (\%). 
Columns (1)–(4) correspond to the input configurations defined in Section \ref{sec:input-con}.}
\footnotesize
\begin{tabular}{|c|c||c|c|c|c|c|}
\hline Train & Test & Model & 1 & 2 & 3 & 4 \\
\hline \multirow{3}{*}{$\mathcal{V}_1$\&$\mathcal{V}_2$}&  \multirow{3}{*}{$\mathcal{V}_3$}& CNN & 24.28 & 60.33 & 74.84 & 76.33  \\
&  & ViT & 26.85 & 68.20 & 73.26 & 78.26 \\
& & LSTM & 27.17 & 74.03 & 80.98 & 84.24\\
& & BiLSTM & 31.96 & 78.26 & 89.22 & 91.11  \\
 &  & BiVTC & 36.09 & 87.50 & 92.76 & 93.85 \\
 & & GF-BiLSTM & - & - & 93.21 & \bf{96.11} \\
\hline
\hline \multirow{3}{*}{$\mathcal{V}_1$\&$\mathcal{V}_3$}&\multirow{3}{*}{$\mathcal{V}_2$} & CNN & 20.57 & 67.93 & 79.31 & 81.65 \\
 & & ViT & 23.04 & 72.28 & 78.80 & 81.52 \\
 & & LSTM & 31.85 & 74.41 & 83.87 & 88.76 \\
& & BiLSTM & 35.81 & 81.52 & 90.85 & 92.93\\
 &  & BiVTC & 41.73 & 86.96 & 93.15 & 95.20\\
 & & GF-BiLSTM & - & - &93.78 & \bf{95.65} \\
\hline
\hline \multirow{3}{*}{$\mathcal{V}_2$\&$\mathcal{V}_3$}&\multirow{3}{*}{$\mathcal{V}_1$}  & CNN & 22.96 & 68.47 & 78.70 & 80.59 \\
 &  & ViT & 25.13 & 68.48 & 73.74 & 79.43 \\
 & & LSTM & 25.36 & 82.70 & 83.91 & 85.84 \\
& & BiLSTM & 33.68 & 80.98 & 86.96 & 88.59 \\
 &  & BiVTC & 35.76 & 84.89 & 91.74 & 94.86 \\
 & & GF-BiLSTM & - & - & 94.33 & \bf{95.10} \\
\hline
\end{tabular}
\label{tab:results}
\end{table}

\begin{table}[t]
\centering
\caption{Preprocessing time (ms) per sample (mean$\pm$sd over 5 runs).}
\vspace{-0.2cm}
\small
\setlength{\tabcolsep}{6pt}
\begin{tabular}{|c|c|c|}
\hline
Preprocessing Method& ms/sample $\downarrow$ & $\times$ vs. Amp \\
\hline
Amp-only (baseline)        & \textbf{1.64} $\pm$ 0.03 & \textbf{1.00} \\
Phase-only (Unw)     & 11.29$\pm$0.11         & 6.88 \\
Amp+Phase (Unw)      & 12.52$\pm$0.07         & 7.65 \\
Phase-only (San)      & 76.71$\pm$0.15         & 46.78 \\
Amp+Phase (San)       & 78.41$\pm$0.27         & 47.89 \\
\hline
\end{tabular}
\vspace{-0.6cm}
\label{tab:pre_time}
\end{table}

\vspace{-0.3cm}
\subsection{Performance Results Analysis}
Table~\ref{tab:results} reports the classification accuracies of all models under the LOVO cross-validation. Phase-only input (Col.~1) is consistently theweakest: even after unwrapping, packet-wise slope/offset and high-frequency jitter leave residual bias and heteroscedastic noise, and without the amplitude, the encoder faces a low-SNR, packet-dependent signal. Amplitude-only (Col.~2) is stronger due to smoother temporal variation. Two-channel inputs (Cols.~3–4) mitigate these issues. Amplitude offers a stable baseline while phase supplies fine timing cues, with sanitized phase (Col.~4) providing a small gain.

The proposed GF-BiLSTM attains the best accuracy on all LOVO splits for dual-channel input. Its robustness stems from reliability-aware fusion and training regularization: a per-time fusion gate adapts the contribution of each stream, and a probabilistic stream-level mask occasionally hides one branch during training. This setup normalizes reliance across modalities, teaching the model to down-weight noisy phase windows and fall back on amplitude when needed, while still exploiting phase when informative—yielding consistent cross-speed gains observed in Table~\ref{tab:results}.

\vspace{-0.3cm}
\subsection{Computational Complexity Analysis}
Using a timing micro-benchmark (I/O excluded), amplitude-only preprocessing costs 1.64\,ms/sample. Adding phase with temporal unwrapping increases cost to 12.52\,ms/sample ($\sim$7.7$\times$), while per-packet linear sanitization raises it to 78.41\,ms/sample ($\sim$47.9$\times$). Both pipelines are $\mathcal{O}(TS)$, but sanitization has a much larger constant due to per-packet regressions. End-to-end, the sanitized pipeline takes 82.08\,ms/sample, with $\sim$79\,ms ($\approx$96\%) spent in the two per-sniffer sanitization steps. In this setting, accuracy gains did not justify the overhead; the unwrap-only variant offers a better accuracy–efficiency trade-off.
\vspace{-0.3cm}
\section{Conclusion} \label{sec:conclusion}
\vspace{-0.3cm}
We systematically studied amplitude and phase representations of Wi-Fi CSI for robotic-arm activity recognition and proposed GF–BiLSTM, a two-stream gated-fusion model. Under the Leave-One-Velocity-Out protocol on RoboFiSense, phase-only was weakest, amplitude-only stronger, and amplitude+phase the best, with sanitized phase providing a consistent gain. GF–BiLSTM achieved the highest accuracy and cross-speed robustness through reliability-aware fusion and light modality dropout. Unwrapped phase offered a better accuracy–efficiency trade-off than linear sanitization. Future directions include polar inputs and complex-valued networks.

\vfill\pagebreak

\bibliographystyle{IEEEbib}
\bibliography{refs}

\end{document}